\documentclass{article}
\usepackage{spconf,amsmath,epsfig,subfigure,multirow,algorithmic,bbding}
\usepackage[ruled]{algorithm2e}
\SetKwRepeat{Do}{do}{while}%
\let\OLDthebibliography\thebibliography
\renewcommand\thebibliography[1]{
  \OLDthebibliography{#1}
  \setlength{\parskip}{0pt}
  \setlength{\itemsep}{0pt plus 0.3ex}
}
\begin{document}\sloppy

\def\x{{\mathbf x}}
\def\L{{\cal L}}

\title{Temporal Up-sampling for Asynchronous Events}

%
\name{Xijie Xiang\textsuperscript{1}\textsuperscript{3}, Lin Zhu\textsuperscript{2}, Jianing Li\textsuperscript{2}, Yonghong Tian\textsuperscript{2}\textsuperscript{3}$^{\ast}$ and Tiejun Huang\textsuperscript{1}\textsuperscript{2}$^{\ast}$\thanks{$^{\ast}$Corresponding author.}}

\address{\textsuperscript{1}Department of Electrical and Computer Engineering, Shanghai Jiao Tong University;\\
\textsuperscript{2}School of Computer Science, Peking University;\\ 
\textsuperscript{3}Peng Cheng National Laboratory.\\
E-mail: ctrump@sjtu.edu.cn, \{linzhu, lijianing, yhtian, tjhuang\}@pku.edu.cn.}

\maketitle


\begin{abstract}

The event camera is a novel bio-inspired vision sensor. When the brightness change exceeds the preset threshold, the sensor generates events asynchronously. The number of valid events directly affects the performance of event-based tasks, such as reconstruction, detection, and recognition. However, when in low-brightness or slow-moving scenes, events are often sparse and accompanied by noise, which poses challenges for event-based tasks. To solve these challenges, we propose an event temporal up-sampling algorithm$\footnote{Code: https://github.com/XIJIE-XIANG/Event-Temporal-Up-sampling}$ to generate more effective and reliable events. The main idea of our algorithm is to generate up-sampling events on the event motion trajectory. First, we estimate the event motion trajectory by contrast maximization algorithm and then up-sampling the events by temporal point processes. Experimental results show that up-sampling events can provide more effective information and improve the performance of downstream tasks, such as improving the quality of reconstructed images and increasing the accuracy of object detection.



\end{abstract}

\vspace{-3mm}

\section{Introduction}
\label{sec:intro}

\vspace{-3mm}

Event cameras (e.g. DVS \cite{DVS}, DAVIS \cite{DAVIS}) are a new type of neuromorphic cameras. Unlike traditional frame-based cameras that generate complete images at fixed time intervals,  event cameras generate data only when it detects brightness changes in the scene. The data of event cameras are spatio-temporal asynchronous events. Each event contains space coordinates $x, y$, generation timestamp $t$ and event polarity $p$ (brightness increase: ON event; brightness decrease: OFF event). This sampling mechanism makes them have the advantages of high temporal resolution, high dynamic range, and low data redundancy. Recently, they have been increasingly used in multimedia community \cite{survey}, such as low-power autonomous driving \cite{jianing} and video reconstruction \cite{zhu1, zhu2,my}.

\begin{figure}[t]
  \centering
  \includegraphics[width=0.85\linewidth]{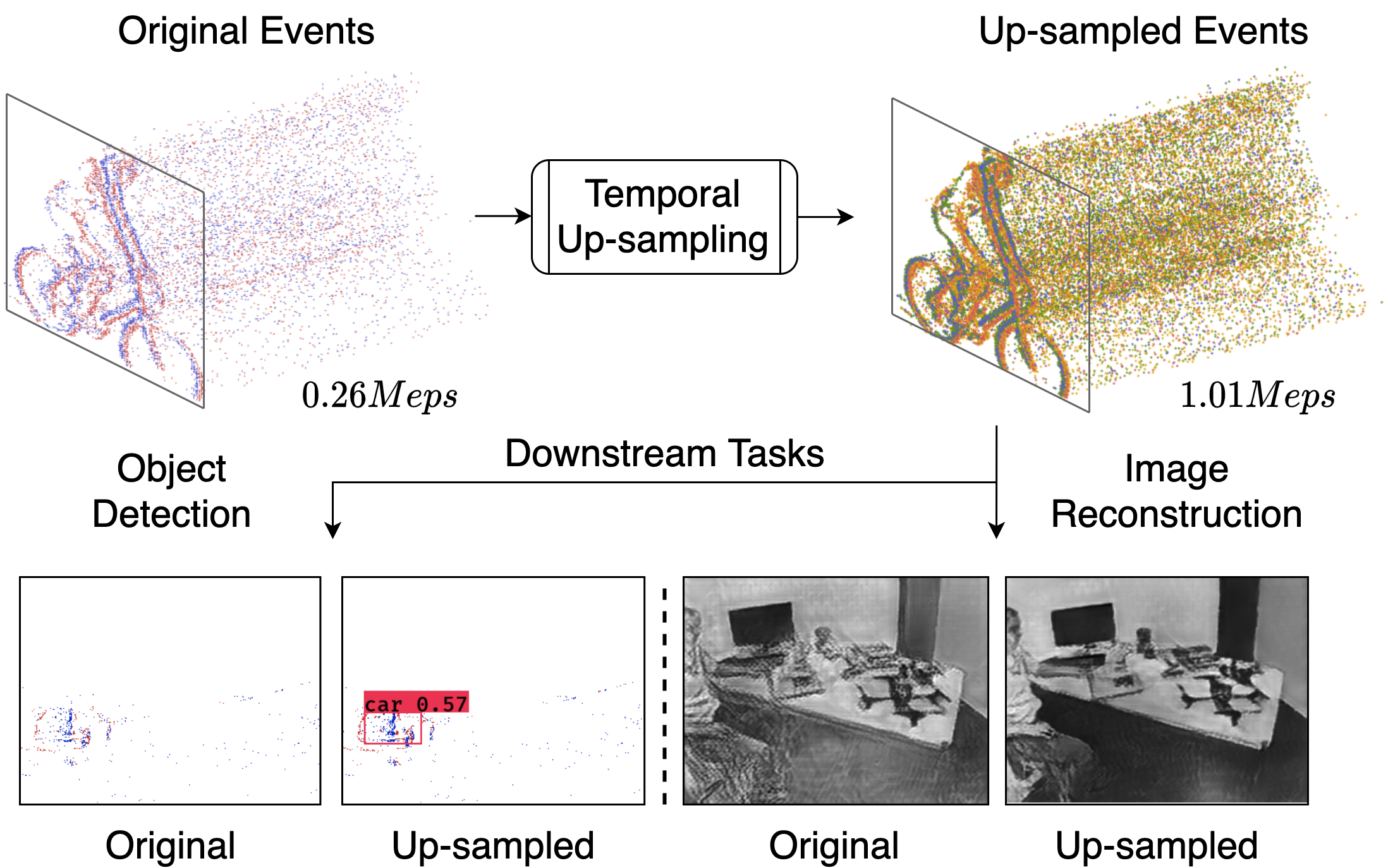}
  \caption{The proposed event temporal up-sampling algorithm, which provides more reliable events for downstream tasks. Red/blue dots: original ON/OFF events, yellow/green dots: up-sampled ON/OFF events. For clarity, all events are accumulated on the beginning time plane.}
  \vspace{-6mm}
  \label{fig:motivation}
\end{figure}

Although event cameras have a high temporal resolution, there are only a few or no events within a small interval in slow-moving or low-brightness scenarios. For slow scenes, the slow movement of objects makes it difficult for the brightness change in a short time to exceed the preset threshold, so only a small number of events are generated. In low-brightness scenes, it is difficult for event cameras to capture dynamic information, and the noise generated is more. Since these scenarios can only provide sparse valid events and contain noise, the performance of event-based tasks is poor. For example, for the image reconstruction task, the limited information provided by sparse events makes the reconstructed image unable to express fine textures and details (original events reconstruction in Fig. \ref{fig:motivation}). For the object detection task, the lack of key information in the original events makes the object difficult to be detected. (detection results of the original events in Fig. \ref{fig:motivation}).

To provide more effective events and suppress noise on event-based tasks, we propose an event temporal up-sampling algorithm for downstream tasks. Our goal is to generate dense up-sampling events from sparse events, which highlights the moving objects or scenes (i.e. main events) and suppresses the noise. To ensure the reliability and accuracy of the up-sampling events, we first estimate the correct motion trajectory of events by contrast maximization algorithm and then distinguish the main events and noise by the number of events on each trajectory, and finally up-sampling the main events and noise through different temporal point processes along the event motion trajectory. 



In general, our contributions are as follows:
\vspace{-\topsep}
\begin{itemize}
\setlength{\itemsep}{0pt}
\setlength{\parsep}{0pt}
\setlength{\parskip}{0pt}
\item We implement event temporal up-sampling along the event motion trajectory, which is consistent with the event generating mechanism of event camera and guarantees the reliability of up-sampling events.
\item In the event up-sampling process, we introduce temporal point processes that consider the impact of historical events to further ensure the reliability of the up-sampling events. In addition, we use different temporal point processes for main events and noise up-sampling respectively to generate more reliable main events and suppress noise up-sampling.
\item The up-sampled events can be used in event-based downstream tasks. Experimentally, up-sampling events increase the number of effective events and improve the performance of downstream tasks, such as increasing the fine texture and details of the reconstructed image and improving the accuracy of object detection.
\vspace{-3mm}
\end{itemize}


\section{Related Work}
\vspace{-3mm}

\textbf{Event spatial up-sampling.} The high temporal resolution of the event camera limits the improvement of its spatial resolution, resulting in the low spatial resolution of the existing event cameras. There are some works \cite{Li2019Super, zoom} to improve the spatial resolution of events, aiming to generate high spatial resolution events from low spatial resolution events. The general process is to first convert asynchronous events into images, then obtain high-resolution images through image-based super-resolution algorithms, and finally randomly generate high-resolution events based on the high-resolution images. Unfortunately, these methods all convert events into images, which weakens the high temporal resolution of the events and makes a limited contribution to the event-based algorithm. In addition, they all use random distribution (e.g. Poisson Process) to generate high-resolution events, and the generated events cannot be guaranteed to be reliable.

\begin{figure*}[htbp]
    \centering
    \includegraphics[width=0.9\linewidth]{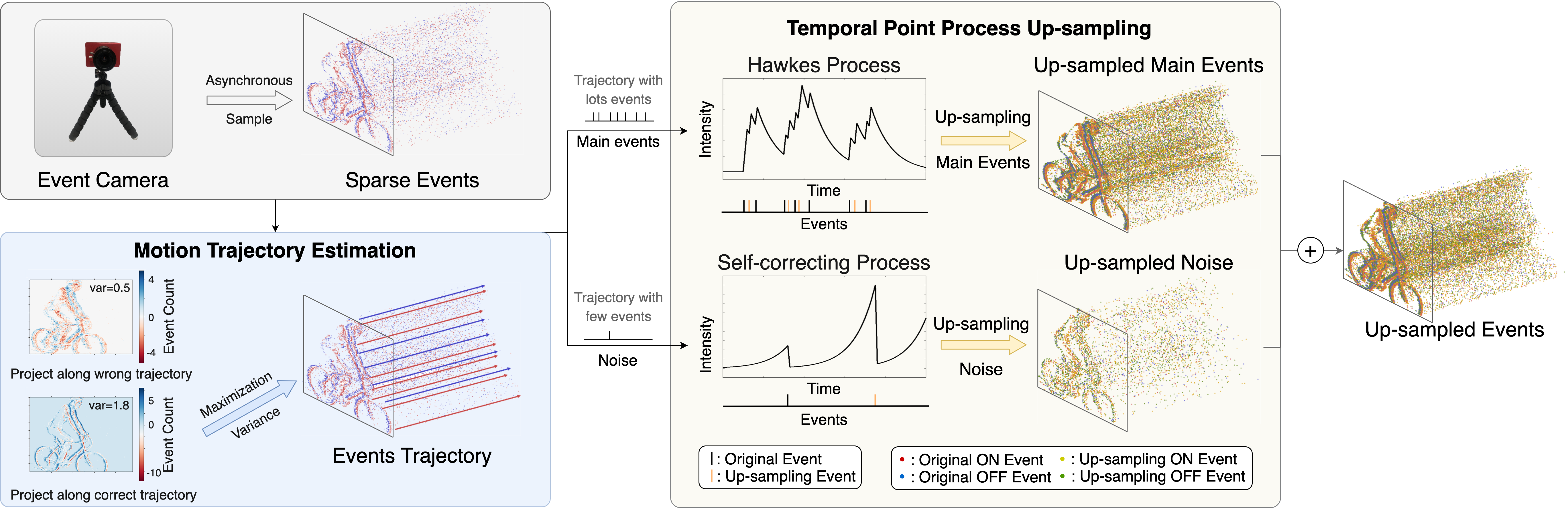}
    \caption{The framework of the proposed event temporal up-sampling algorithm. The input data is sparse events collected asynchronously by the event camera. First, the correct motion trajectory is estimated by contrast (i.e. variance) maximizing algorithm. Then, for two different types of events, two point process models are used to generate new up-sampling events on the estimated trajectory. Finally, the two up-sampling events are combined (\textcircled{+}) into the final up-sampling result. For clarity, all events are accumulated on the plane at the start time.}
    \vspace{-5mm}
    \label{fig:framework}
\end{figure*}

\textbf{Super-resolution reconstruction of events.}
Another type of related work, event super-resolution reconstruction, combines event up-sampling and image reconstruction tasks together. It aims to generate high-resolution reconstructed images from low-resolution events. Mohammad et al. \cite{choi2020learning} expressed events as an accumulated spike number image over a period of time and then uses CNNs to achieve super-resolution reconstruction. Wang et al. \cite{wang2020eventsr} converted spikes into intensity images and then used a 3-phase GANs for super-resolution reconstruction. In these related work, event up-sampling can improve the performance of downstream tasks, such as object tracking \cite{zoom} and reconstruction \cite{wang2020eventsr}, which proves the event up-sampling is meaningful.

\section{Method}


\subsection{Event Temporal Up-sampling}
\label{ETU}
The purpose of event temporal up-sampling is to generate dense up-sampled events $\mathcal{E}_d=\{e_j\}_{j=1}^N$ from sparse events $\mathcal{E}_s=\{e_i\}_{i=1}^M$, where $N, M$ are the number of dense events and sparse events, $N>M$. Recall that each event $e=\{x, y, p, t\}$. We aim to generate dense events $\mathcal{E}_d$ along the motion trajectory of sparse events $\mathcal{E}_s$:

\begin{equation}
    \mathcal{E}_d = \mathcal{G}(\mathcal{E}_s; \theta^*),
\end{equation}

\noindent where $\theta^*$ is the estimated motion trajectory of sparse events, $\mathcal{G}$ is the generating operation for up-sampling. The complete up-sampling process is described in Algorithm \ref{algo}.



\begin{algorithm}
  \SetKwInOut{KIN}{Input}
  \SetKwInOut{KOUT}{Output}
  \KIN{Sparse events: $\mathcal{E}_s=\{e_i\}_{i=1}^M$}
  
  \KOUT{Dense up-sampled events: $\mathcal{E}_d=\{\mathcal{E}_d^k|k=1,2,...\} $}
  
  \textbf{Step 1:} Estimate Motion Trajectory $\theta^*$
  
  \Do{$f(\theta)<\mathrm{max}(f(\theta))$}{
    Initialize a random trajectory $\theta=(v_x,v_y)$;
    
    \For{\rm{each} $e_i, i \in [1,M]$}{
    
    Warp $e_i$ to reference time plane $t_{ref}$ along trajectory $\theta$: $e_i' \leftarrow e_i$;
    }
    Calculate the events number: $H(x,y;\theta)$;
    
    Calculate the variance: $f(\theta)=\mathrm{var}(H(x, y;\theta))$;
    
    }
  \textbf{Step 2:} Temporal Point Process Up-sampling
  
  \For{ \rm{each trajectory} $tr_k,  k=1,2,...$}{
  
    \eIf{\rm{Events number in} $tr_k$ \rm{is greater than threshold} $\phi$}{
    
    Up-sampling events by Hawkes Process:
    
    $\mathcal{E}_d^k=\mathcal{G}_\mathrm{HP}(\mathcal{E}_s^k;\theta^*)$
    }{
    Up-sampling events by Self-correcting Process:
    
    $\mathcal{E}_d^k=\mathcal{G}_\mathrm{SP}(\mathcal{E}_s^k;\theta^*)$
    }
    }
  \caption{Event Temporal Up-sampling}
  \label{algo}
\end{algorithm}


Since the event camera is sensitive to the dynamic information in the scene, the event will only be generated when the change in brightness exceeds the threshold. The visual effect presented is that sparse events roughly depict the edge of moving objects (cumulative events image at the start time in the gray box of Fig. \ref{fig:framework}). Over time, new events are generated along the trajectory of objects moving relative to the camera. To ensure the accuracy of the up-sampling events and conform to the sampling principle of the event camera, the up-sampling events should be generated on the object motion trajectory. Therefore, the problem turns into how to estimate the correct trajectory of moving objects (Sec. \ref{MTE}).

Then, we up-sample new events on the correct motion trajectory. Some existing work \cite{Li2019Super, zoom} randomly generate events through Poisson Process. But Poisson Process is memoryless and cannot guarantee the consistency of newly generated events with the original events. To have a basis for event up-sampling, we introduce temporal point processes (Sec. \ref{TPPG}) with memory, including Hawkes Process \cite{HP} and Self-Correcting Process \cite{SCP}, which consider the impact of historical events.

\subsection{Motion Trajectory Estimation}
\label{MTE}

The contrast maximization algorithm \cite{CM} can effectively estimate the trajectory of events, and has been successfully used in tasks such as denoising \cite{9156457} and motion estimation \cite{9329204}. The main idea of contrast maximization is to warp events to a certain time plane along the estimated motion trajectory and count the number of events in the time plane. When the variance (i.e. contrast) of the number of these events is the largest, the estimated motion trajectory is correct. As shown in the blue box of Fig. \ref{fig:framework}, the edges projected on the reference time plane along the wrong trajectory are blurred and the variance is small. This is due to the wide range of pixels in the event distribution after projecting. On the contrary, along the correct trajectory projection, a large number of events after projection are concentrated on a small number of pixels, and the variance is large. The goal of this algorithm is to maximize the contrast To find the best motion trajectory.


Specifically, we first initialize a random motion trajectory $\theta=(v_x, v_y)$ and warp events along this trajectory to a reference time $t_{ref}$. The coordinate $(x_i', y_i')$ of warped event $e_i'$ are 

\begin{equation}
    (x_i', y_i') = (x_i, y_i) + (t_{ref} - t_i)\theta,
\end{equation}

\noindent where $x_i, y_i, t_i$ is the coordinate and timestamp of original event $e_i$, the warped event maintains the same polarity $p_i$ as the original event. We choose the timestamp of the last event as the reference time.

Then we count the number of events for each pixel $(x, y)$ on the reference time plane, 

\begin{equation}
    H(x,y;\theta)=\sum_{i=1}^M p_i' \delta ((x, y)-(x_i',y_i')),
\end{equation}

\noindent where $\delta(\cdot)$ is Dirac function. Only when the warped event $e_i'=\{x_i', y_i', p_i', t_i'\}$ falls into the current coordinate $(x,y)$, this warped event $e_i'$ at $(x,y)$ is recorded.

Finally, we calculate the variance of events number: $f(\theta) = \mathrm{var}(H(x,y;\theta))$. The objective of motion trajectory estimation is as follow:

\begin{equation}
    \theta^*=\mathrm{argmax}_\theta f(\theta).
\end{equation}

We use the Nelder-Mead algorithm \cite{NM} to optimize the parameter $\theta$. When the variance gradually converges to the maximum value, the optimal motion trajectory $\theta^*$ are obtained. It will be used in the temporal up-sampling events generation.

\subsection{Temporal Point Process Up-sampling}
\label{TPPG}
To generate more reliable up-sampling events, we introduce temporal point processes to generate events on the estimated motion trajectory. In the motion trajectories, there are a large number of events in some trajectories, which are often triggered by the edges of moving objects in the scene. We classify this type of events as the main events. While some trajectories have only a few discrete events, and these points are considered noise. For these two types of events, we use two point processes (i.e. Hawkes Process \cite{HP} and Self-Correcting Process \cite{SCP}) modeling respectively (yellow box of Fig. \ref{fig:framework}).

\textbf{Hawkes Process.} To generate more up-sampling events on the trajectories of the edges of the moving object, we use Hawkes Process to implement event up-sampling of the main events and use Ogata's modified thinning algorithm \cite{Ogata} to simulate Hawkes Process. Its main idea is that each historical event has a positive impact to increase the generation probability of the current event (Hawkes Process in the yellow box of Fig. \ref{fig:framework}). Besides, the farther the historical event is from the current moment, the smaller the impact. As follows,

\begin{equation}
    \lambda_H^*(t)=\mu+\alpha\sum_{t_i<t}\mathrm{exp}(-(t-t_i)),
\end{equation}

\noindent where $\lambda_H^*(t)$ is the conditional intensity function of Hawkes Process. The greater the intensity, the greater the probability of generating an event. $\mu$ is the basic intensity of original sparse events, $\alpha$ and the time difference between the historical event $t_i$ and the current moment $t$ together determine the impact from each historical event. Through the Hawkes Process, we can generate more dense up-sampling events based on the impact of historical events.


\textbf{Self-correcting Process.} We use self-correcting process to simulate the up-sampling of noise. Over time, the probability of noise increases. But when a noise is generated or there is an original noise in current time, the probability of this generation drops by $\mathrm{exp}(\beta)$ immediately (Self-correcting Process in the yellow box of Fig. \ref{fig:framework}). The intensity function is

\begin{equation}
    \lambda_S^*(t)=\mathrm{exp}(\mu t-\sum_{t_i<t} \beta).
\end{equation}


\noindent Self-correcting Process can control the generation number of noise as little as possible.

Finally, the up-sampling results of the two point processes (Up-sampling events by Hawkes Process: $\mathcal{G}_{HP}$, Up-sampling events by Self-correcting Process: $\mathcal{G}_{SP}$) are merged \textcircled{+} into complete up-sampling events,

\begin{equation}
    \mathcal{E}_d=\mathcal{G}_{HP}(\mathcal{E}_s;\theta^*) \textcircled{+} \mathcal{G}_{SP}(\mathcal{E}_s;\theta^*).
\end{equation}

\subsection{Computational Complexity}
For the motion trajectory estimation, the computation mainly comes from the event warping (for loop of step 1 in Algorithm \ref{algo}) and parameter optimization (step 1 in Algorithm \ref{algo}). The computational complexity of event warping is linear on the event number. And using the nonshrink iteration of the Nelder-Mead algorithm, only one or two test points need to be calculated for each iteration. The time complexity is $O(n)$.

For the temporal point process up-sampling, the main calculation lies in the simulation of the point process. Similar to the previous step, we also adopt an iterative approach. We record the impact of previous historical events and only the latest historical events need to be considered in each iteration. The time complexity is also $O(n)$.

\section{Experiments}
\subsection{Consistency Verification}
\label{exp1}
\begin{figure}[htbp]
  \centering
  \subfigure[Numbers of original and up-sampled ON events.]{
  \begin{minipage}[t]{0.45\linewidth}
  \centering
  \includegraphics[width=\linewidth]{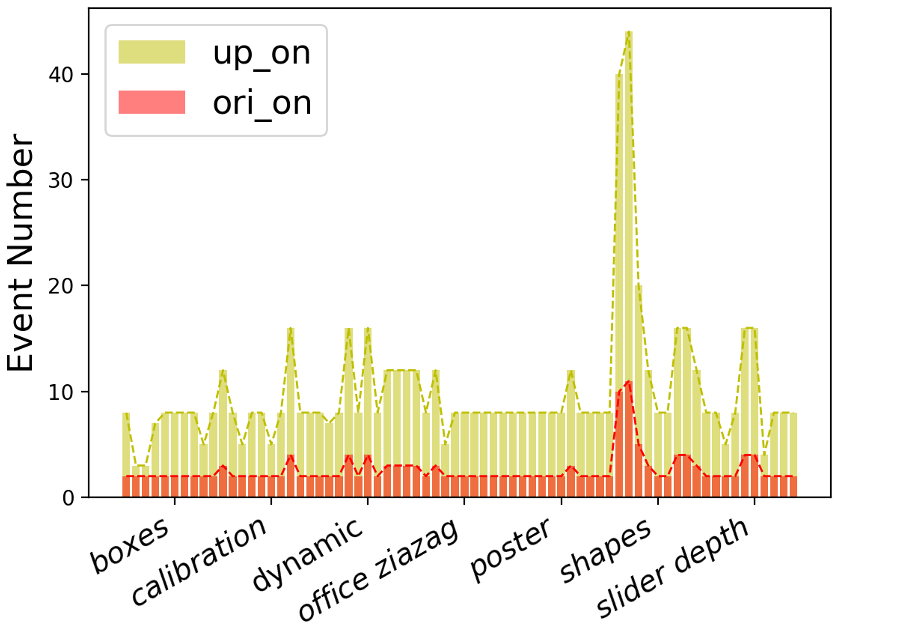}
  \end{minipage}%
  }%
  \centering
  \subfigure[Numbers of original and up-sampled OFF events.]{
  \begin{minipage}[t]{0.45\linewidth}
  \centering
  \includegraphics[width=\linewidth]{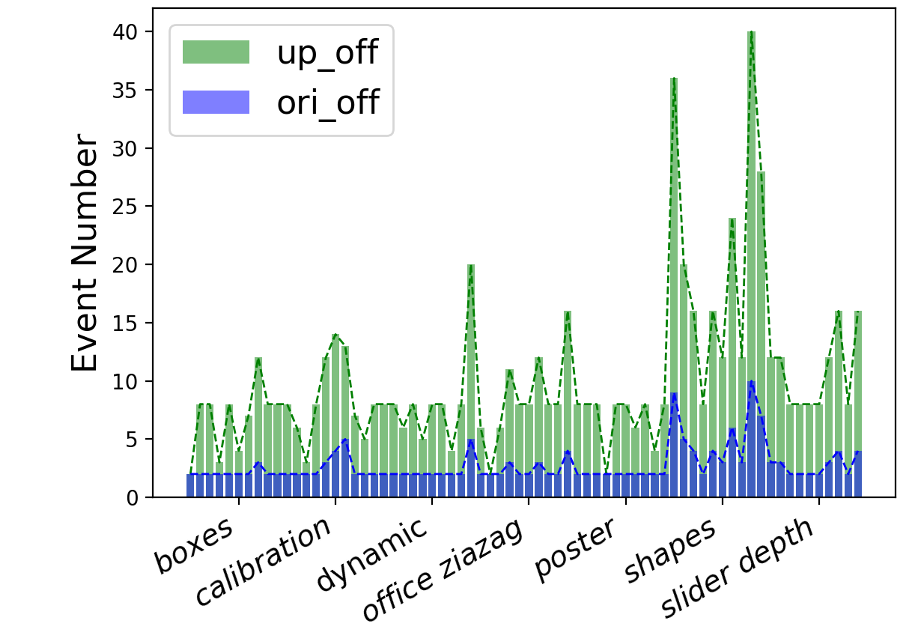}
  \end{minipage}%
  }%
  \centering
  \caption{Comparison between original and up-sampled event number events for each trajectory.}
  \label{event_trajectory}
\end{figure}

The up-sampled events should be consistent with the original event, rather than randomly generated. To verify this consistency, we randomly select ten event trajectories from each of the seven scenarios in the IJRR dataset \cite{IJRR}, and count the number of original events and up-sampled events. As shown in Fig. \ref{event_trajectory}, the event intensity after up-sampling is mostly stronger than the original event intensity. A trajectory with a larger number of original events tends to generate more up-sampled events, which matches our expectations.

\subsection{Ablation Experiments}
\label{exp2}
To verify the influence of the motion trajectory on the up-sampling result, we use $\theta=(0,0)$ to replace the trajectory estimated by the contrast maximization (CM). In addition, we replaced Hawkes Process and Self-correcting Process with Poisson Process to verify the influence of the temporal point processes (TPPs). We evaluate the variance and gradient of the projected event image along the corresponding trajectory of the original events and the up-sampling events on IJRR dataset. The greater the variance, the richer the information that the events can represent. The larger the gradient, the sharper the edge of the event image. As shown in Table \ref{table:CMTPP}, replacing CM and TPPs damage the effect of up-sampling. It proves that up-sampling along the event motion trajectory through temporal point processes is effective.


\begin{table}[htbp]
\footnotesize
\centering
\caption{\textbf{Ablation Experiment}}
\label{table:CMTPP}
\begin{tabular}{cccc}
\hline
\textbf{CM} & \textbf{TPPs} & \textbf{Variance ($\uparrow$)} & \textbf{Gradient ($\uparrow$)} \\ \hline
\textbf{\XSolidBrush}                      & \textbf{\Checkmark}                       & 1.33              & 0.34              \\
\textbf{\Checkmark}                      & \textbf{\XSolidBrush}                       & 2.21              & 0.56              \\ \hline
\textbf{\Checkmark}                      & \textbf{\Checkmark}                       & \textbf{6.49}     & \textbf{0.68}     \\ \hline
\end{tabular}
\end{table}


\begin{figure*}[htbp]
    \centering
    \includegraphics[width=1\linewidth]{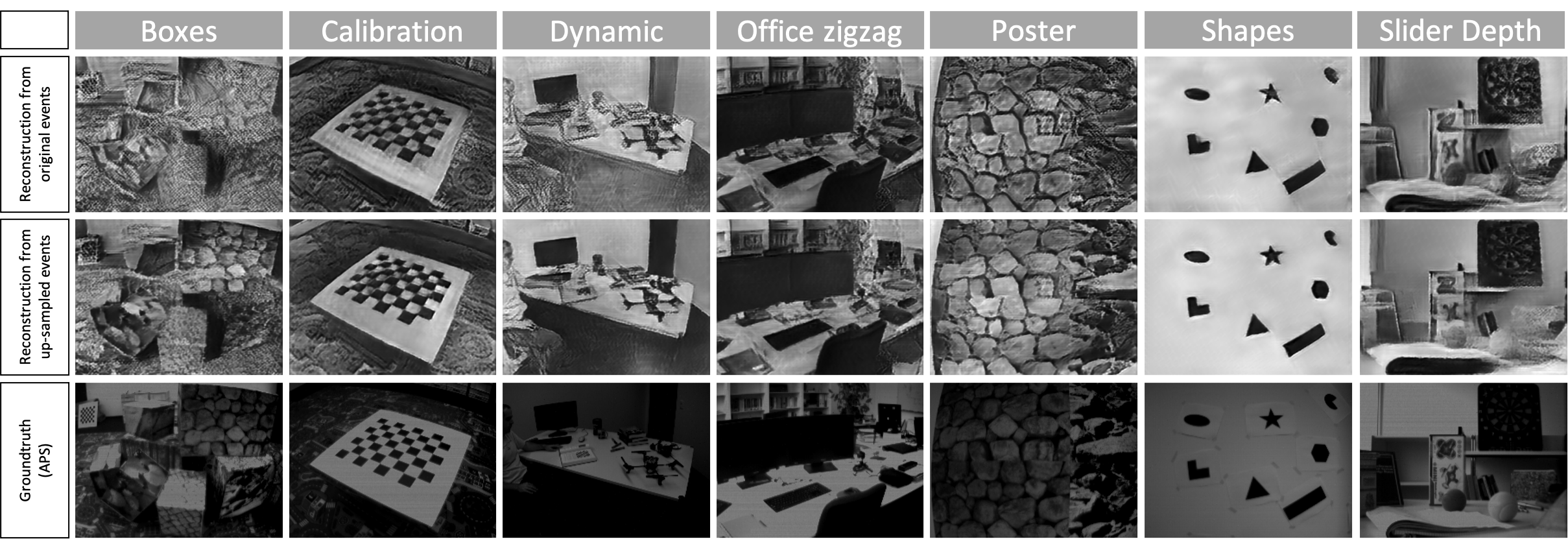}
    \caption{Reconstructed results on IJRR dataset.}
    \label{fig:recon}
\end{figure*}

\subsection{Up-sampling for Image Reconstruction}
\label{exp3}
We perform image reconstruction experiments on the IJRR dataset. In each scene of the dataset, we select events of about 2.2 seconds (minimum number of events: 153,679; maximum number of events: 1,438,276), corresponding to 50 frames of groundtruth. We first up-sample the original events through our algorithm, and then reconstruct the original and the up-sampled events of $1ms$ into the images respectively by reconstruction algorithm \cite{recon1}. Although the inputs (the original events and the up-sampling events) are in the same time period, the number of up-sampled events is larger and the effective information carried is more. Therefore, the effect of events reconstruction after up-sampling is better. This advantage is more obvious when the number of original events is smaller and the accumulation time is shorter (Appendix 2).

\begin{table}[htbp]
\footnotesize
\centering
\caption{\textbf{Quantitative evaluation on IJRR dataset}}
\label{table:recon}
\begin{tabular}{c|cc|cc|cc}
\hline
\multirow{2}{*}{\textbf{Scene}} & \multicolumn{2}{c|}{\textbf{PL ($\downarrow$)}} & \multicolumn{2}{c|}{\textbf{MSE ($\downarrow$)}} & \multicolumn{2}{c}{\textbf{SSIM ($\uparrow$)}} \\ \cline{2-7} 
                                & \textbf{Ori}    & \textbf{Up}    & \textbf{Ori}    & \textbf{Up}     & \textbf{Ori}    & \textbf{Up}     \\ \hline
\textbf{Boxes}                  & 0.42            & \textbf{0.35}  & 0.10            & \textbf{0.08}   & 0.27            & \textbf{0.33}   \\
\textbf{Calibration}            & 0.48            & \textbf{0.42}  & 0.08            & \textbf{0.07}   & 0.32            & \textbf{0.38}   \\
\textbf{Dynamic}                & 0.50            & \textbf{0.42}  & 0.07            & \textbf{0.06}   & 0.41            & \textbf{0.53}   \\
\textbf{Office zigzag}          & 0.43            & \textbf{0.35}  & 0.05            & \textbf{0.04}   & 0.55            & \textbf{0.60}   \\
\textbf{Poster}                 & 0.28            & \textbf{0.24}  & 0.06            & \textbf{0.05}   & 0.49            & \textbf{0.54}   \\
\textbf{Shapes}                 & 0.50            & \textbf{0.49}  & 0.17            & \textbf{0.15}   & \textbf{0.43}   & 0.40            \\
\textbf{Slider depth}           & \textbf{0.37}   & \textbf{0.37}  & 0.06            & \textbf{0.05}   & 0.55            & \textbf{0.57}   \\ \hline
\textbf{Average}                & 0.43            & \textbf{0.38}  & 0.08            & \textbf{0.07}   & 0.43            & \textbf{0.48}   \\ \hline
\end{tabular}
\end{table}


In Table \ref{table:recon}, we use three evaluation indicators to evaluate the similarity between the reconstructed images and groundtruth, i.e. perceptual loss (PL), mean square error (MSE), and structural similarity (SSIM), and the best performing results are shown in bold. The groundtruth comes from the active pixel sensor (APS) of event camera, DAVIS. Lower PL, MSE and higher SSIM means better results. The reconstruction results of up-sampling events perform better than the original events in almost all scenes, which proves that up-sampling provide more effective events for reconstruction tasks and help improve the quality of reconstructed images. For the shape scene, the SSIM decreases slightly. A possible reason is that the scene is composed of simple shapes, and the reconstruction result of the original events already has a high structural similarity with the groundtruth. However, the qualitative results in Fig. \ref{fig:recon} (6th column) show that the reconstructed shapes of the up-sampling events are more complete and clear. In addition, the reconstruction algorithm continuously accumulates previous information. Since the up-sampling events have more effective information in the early stage, a good reconstruction effect can be achieved faster.




\begin{figure*}[htbp]
    \centering
    \includegraphics[width=1\linewidth]{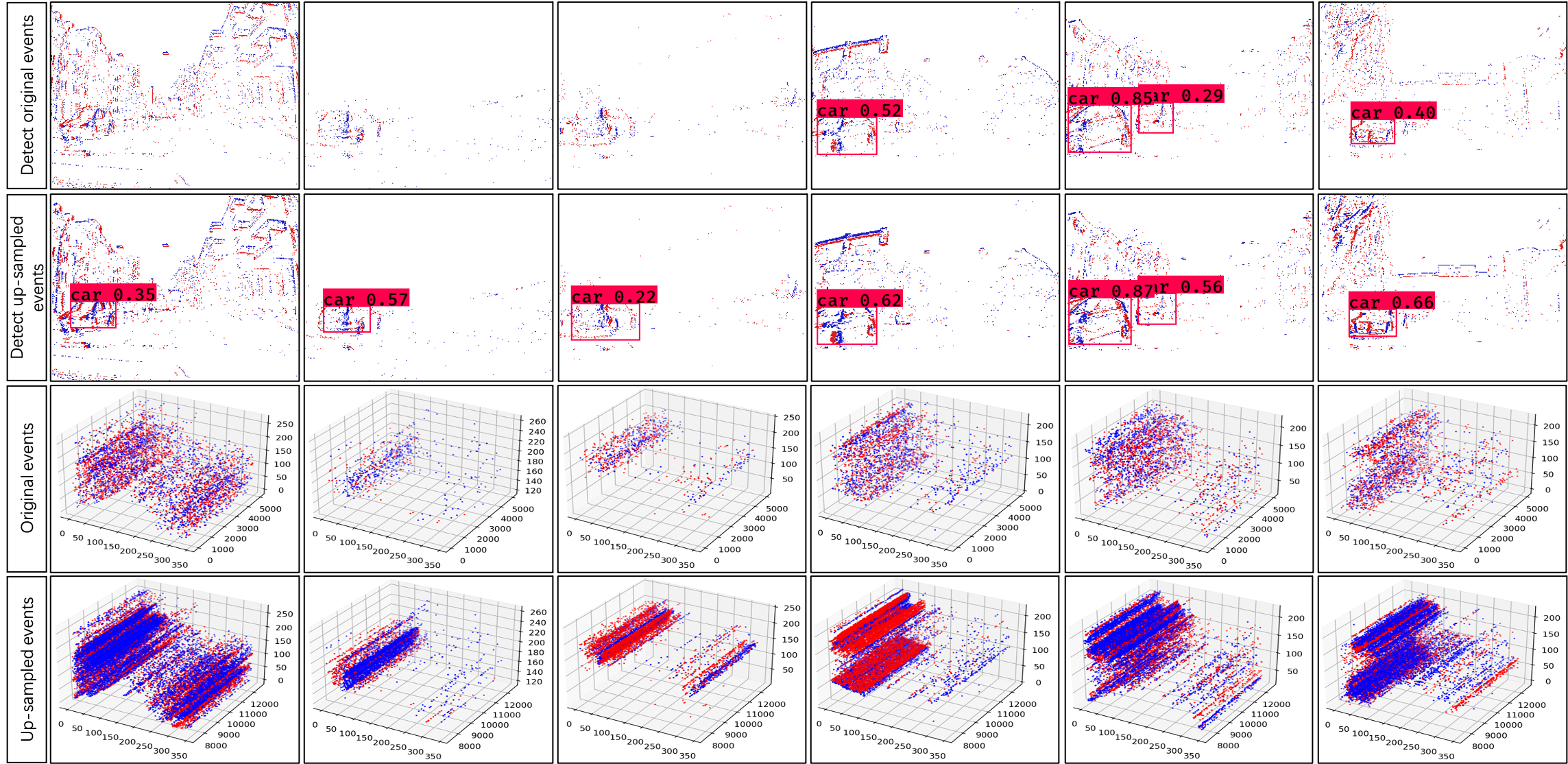}
    \caption{Detection results on DDD17-PKU-CAR dataset. Red dots indicate ON events, blue dots indicate OFF events}
    \label{fig:detect}
\end{figure*}

\subsection{Up-sampling for Object Detection}
\label{exp4}
We perform object detection experiments on DDD17-PKU-CAR dataset \cite{jianing}. The dataset contains daytime and night scenes, covering normal events and low-brightness conditions where the events are sparse and noisy. We up-sample the test set and use the detection model proposed in \cite{jianing} to detect.

To simulate different event sparsity levels, we select different time intervals (5ms, 7ms and 10ms). The detection rates of original events are 0.13, 0.19, and 0.25. The event detection rates after up-sampling are 0.16, 0.20, 0.27. Experimental results show that the longer the time, the higher the detection rate of the original events and the up-sampling events. In addition, in all selected time intervals, the event detection rate after up-sampling is higher than that of the original event, which proves that the proposed up-sampling algorithm can provide more effective information for the detection task.

Fig. \ref{fig:detect} shows the qualitative detection results. As show in the first three columns of Fig. \ref{fig:detect}, the car outline of original events is not clear, so the car cannot be detected. After up-sampling, the number of events expressing the car is large enough to be detected. As show in the last three columns of Fig. \ref{fig:detect}, the target car can be detected from the original events, but the higher the confidence of detecting after up-sampling. In addition, to further compare the difference between the up-sampling and the original events, we show the spatial-temporal distribution of them in the last two rows of Fig. \ref{fig:detect}. It can be seen that the up-sampling and the original events are basically on the same trajectory, but the up-sampling events are more dense and with more information.


\section{Conclusion}


In this paper, we present a novel event-based task, namely event temporal up-sampling, which aims to up-sample high temporal resolution events from asynchronous and sparse events. We first estimate events motion trajectories by contrast maximization algorithm and then generate up-sampling events by two different temporal point processes along the estimated motion trajectory. Experimentally, the up-sampled events can achieve better image reconstruction quality and higher object detection accuracy. We believe that this work will improve the advantages of the event camera on various vision tasks in the future.


\section{ACKNOWLEDGEMENT}


This work was supported in part by the National Natural Science Foundation of China under Contract 62027804, Contract 61825101 and Contract 62088102.

\begin{figure*}[bp]
    \centering
    \includegraphics[width=1\linewidth]{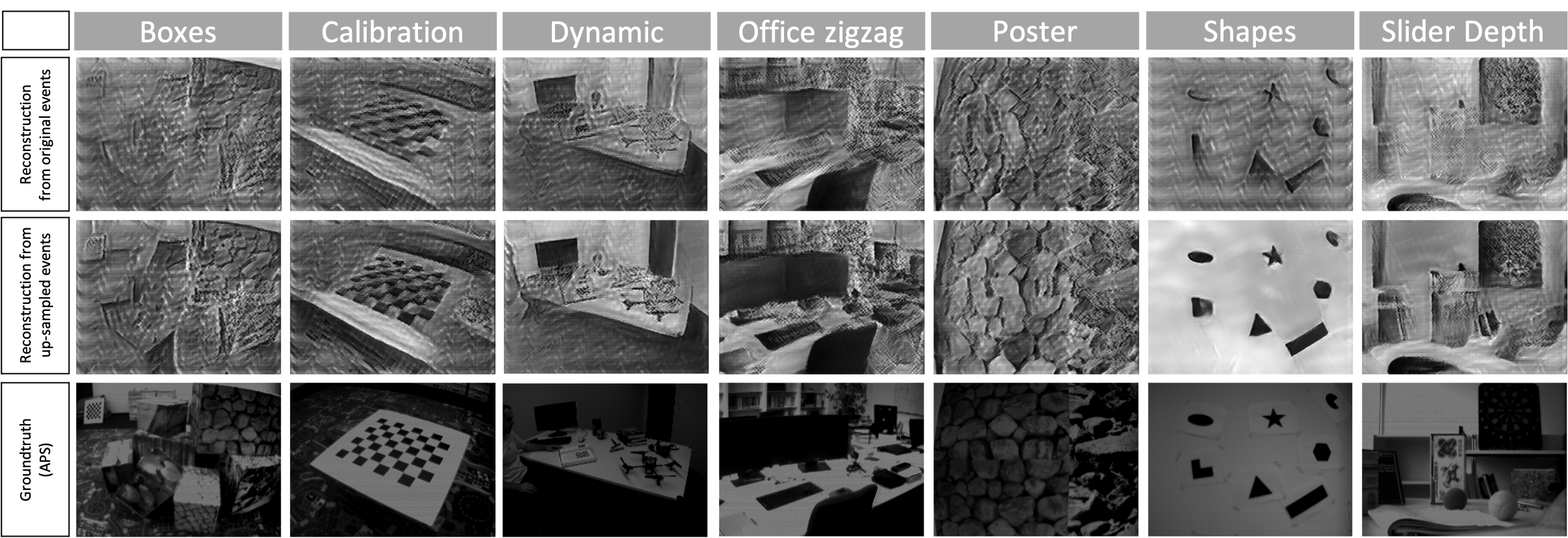}
    \caption{Reconstructed results on IJRR dataset.}
    \label{fig:recon2}
\end{figure*}

\begin{figure*}[htbp]
    \centering
    \includegraphics[width=1\linewidth]{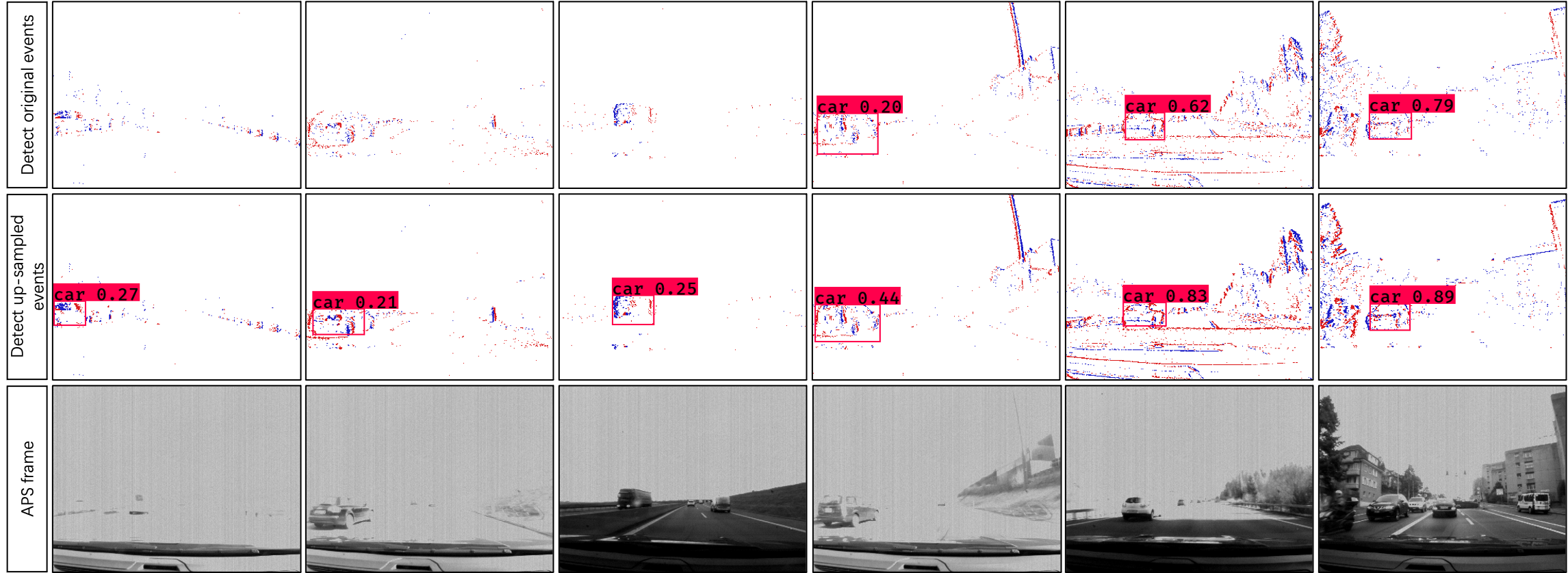}
    \caption{Detection results on DDD17-PKU-CAR dataset \cite{jianing}. Red dots indicate ON events, blue dots indicate OFF events}
    \label{fig:detect2}
\end{figure*}

\section{Appendix}
\subsection{Parameters Setting}

\textbf{Parameters setting of Hawkes Process.} Given $N$ events in a period of time $T$, they are located on $k$ trajectories. The goal of Hawkes Process is to generate up-sampled events for the trajectory where the main event is located. According to the simulate algorithm of Hawkes Process (Ogata’s modified thinning algorithm \cite{Ogata}), the interval $\Delta t$ between the time of the next event and the current time obeys the exponential distribution of intensity $\lambda$, 

\begin{equation}
    \Delta t \sim \mathrm{Exp}(\lambda).
\end{equation}

\noindent The intensity $\lambda$ is affected by three factors, the number of events on the trajectory, the ratio of main events to noise, and the minimum interval between events on the trajectory. More specifically, we specify a base intensity $\lambda_0=\frac{N/k}{T}$ (average intensity of input events) so that the generated time interval is within a reasonable range. Then, we add the influence of the three factors to the basic intensity to obtain the intensity of each trajectory,

\begin{equation}
    \lambda=\frac{\lambda_0 \times n_k \times (\frac{n_{m}^{ori}}{n_{n}^{ori}} + r)}{\log(\Delta t_{min})},
\end{equation}

where $n_k, n_{m}^{ori}, n_{n}^{ori}$ are the number of input events on $k$th trajectory, the number of main events and the number of noise. $r$ is the up-sampling rate. $(\frac{n_{m}^{ori}}{n_{n}^{ori}} + r)$ is derived from $n_{m}^{up}+n_{n}^{up} = (n_{m}^{ori}+n_{n}^{ori}) \times r$.

\textbf{Parameters setting of Self-correcting Process.} We use Self-correcting Process to up-sample events for trajectories with only one event (i.e. noise). Recall that the probability of event generation increases $\mathrm{exp}(\mu t)$ over time. $\mu$ is the base intensity, which is the same as the base intensity in Hawkes Process. Since there is only one event for each trajectory, the basic intensity is very low, ensuring that noise will not be over-generated.

\begin{table}[htbp]
\centering
\caption{\textbf{Quantitative evaluation on IJRR \cite{IJRR} dataset}}
\label{table:recon2}
\begin{tabular}{c|cc|cc|cc}
\hline
\multirow{2}{*}{\textbf{Scene}} & \multicolumn{2}{c|}{\textbf{PL($\downarrow$)}}         & \multicolumn{2}{c|}{\textbf{MSE($\downarrow$)}}        & \multicolumn{2}{c}{\textbf{SSIM($\uparrow$)}}        \\ \cline{2-7} 
                                & \textbf{Ori} & \textbf{Up} & \textbf{Ori} & \textbf{Up} & \textbf{Ori} & \textbf{Up} \\ \hline
\textbf{Boxes}                  & 0.59            & \textbf{0.49}      & 0.14            & \textbf{0.12}      & 0.18            & \textbf{0.20}      \\
\textbf{Calibration}            & 0.52            & \textbf{0.50}      & 0.14            & \textbf{0.13}      & 0.16            & \textbf{0.19}      \\
\textbf{Dynamic}                & 0.63            & \textbf{0.59}      & 0.10            & \textbf{0.09}      & 0.26            & \textbf{0.30}      \\
\textbf{Office zigzag}          & 0.72            & \textbf{0.57}      & 0.18            & \textbf{0.10}      & 0.23            & \textbf{0.34}      \\
\textbf{Poster}                 & 0.56            & \textbf{0.48}      & 0.12            & \textbf{0.10}      & 0.24            & \textbf{0.29}      \\
\textbf{Shapes}                 & 0.54            & \textbf{0.53}      & 0.15            & \textbf{0.14}      & 0.21            & \textbf{0.31}      \\
\textbf{Slider depth}           & 0.55            & \textbf{0.50}      & 0.09            & \textbf{0.07}      & 0.30            & \textbf{0.37}      \\ \hline
\textbf{Average}                & 0.59            & \textbf{0.52}      & 0.12            & \textbf{0.11}      & 0.23            & \textbf{0.29}      \\ \hline
\end{tabular}
\end{table}


\subsection{More Results of Image Reconstruction}
We select events with a time interval of 100us to reconstruct the image to simulate a more extreme sparse scene. The qualitative and quantitative results are shown in the Fig. \ref{fig:recon2} and Table. \ref{table:recon2} respectively. It can be seen from the results that the overall reconstruction effect has declined, which is caused by the fact that sparse events can only provide very little effective information. But up-sampling can still improve the quality of reconstructed images even for such sparse input events.

\subsection{More Results of Vehicle Detection}
Fig. \ref{fig:detect2} shows the detection result and the corresponding APS frame. Up-sampling events can improve detection accuracy (the first three columns in Fig. \ref{fig:detect2}) or the confidence of detection (the last three columns in Fig. \ref{fig:detect2}).

\small
\bibliographystyle{IEEEbib}
\bibliography{icme2022template}

\end{document}